\documentclass{article}

\usepackage{arxiv}

\usepackage[utf8]{inputenc} % allow utf-8 input
\usepackage[T1]{fontenc}    % use 8-bit T1 fonts
\usepackage{hyperref}       % hyperlinks
\usepackage{url}            % simple URL typesetting
\usepackage{booktabs}       % professional-quality tables
\usepackage{amsfonts}       % blackboard math symbols
\usepackage{nicefrac}       % compact symbols for 1/2, etc.
\usepackage{microtype}      % microtypography
\usepackage{lipsum}
\usepackage{graphicx}
\usepackage{subfigure}
\graphicspath{ {./images/} }

\title{LeViT-UNet: Make Faster Encoders with Transformer for Medical Image Segmentation}

\author{
 Guoping Xu \\
  School of Computer Sciences and Engineering\\
  Wuhan Institute of Technology\\
  Wuhan, Hubei, China, 430205 \\
  \texttt{xugp2006@126.com} \\
   \And
 Xingrong Wu{*} \\
  School of Computer Sciences and Engineering\\
  Wuhan Institute of Technology\\
  Wuhan, Hubei, China, 430205 \\
  \texttt{xwu@wit.edu.cn} \\
  
     \And
 Xuan Zhang \\
  School of Computer Sciences and Engineering\\
  Wuhan Institute of Technology\\
  Wuhan, Hubei, China, 430205 \\
  
     \And
  Xinwei He \\
  School of Electronic Information and Communications\\
  Huazhong University of Science and technology\\
  Wuhan, Hubei, China, 430074 \\ 
  
}

\begin{document}
\maketitle
\begin{abstract}
Medical image segmentation plays an essential role in developing computer-assisted diagnosis and therapy systems, yet still faces many challenges. In the past few years, the popular encoder-decoder architectures based on CNNs (e.g., U-Net) have been successfully applied in the task of medical image segmentation. However, due to the locality of convolution operations, they demonstrate limitations in learning global context and long-range spatial relations. Recently, several researchers try to introduce transformers to both the encoder and decoder components with promising results, but the efficiency requires further improvement due to the high computational complexity of transformers. 
In this paper, we propose LeViT-UNet, which integrates a LeViT Transformer module into the U-Net architecture, for fast and accurate medical image segmentation. Specifically, we use LeViT as the encoder of the LeViT-UNet, which better trades off the accuracy and efficiency of the Transformer block. Moreover, multi-scale feature maps from transformer blocks and convolutional blocks of LeViT are passed into the decoder via skip-connection, which can effectively reuse the spatial information of the feature maps. Our experiments indicate that the proposed LeViT-UNet achieves better performance comparing to various competing methods on several challenging medical image segmentation benchmarks including Synapse and ACDC. Code and models will be publicly available at https://github.com/apple1986/LeViT\_UNet.

\end{abstract}

% keywords can be removed
%\keywords{Segmentation \and Transformer \and UNet}

\section{Introduction}
%\paragraph{} 
Automated medical image segmentation has been widely studied in the medical image analysis community which would significantly reduce the amount of tedious and error-prone work by radiologists. In the past few years, Convolutional Neural Networks (CNNs) have made substantial progress in medical image segmentation. Fully convolutional networks (FCNs)\cite{1fcn} and its variants (e.g., U-Net\cite{2unet}, SegNet\cite{3segnet}, DeepLab\cite{4deeplab}, CCNet\cite{5ccnet}) are extensively used architectures. They have been applied in cardiac segmentation from MRI\cite{6bpd}, liver and tumor segmentation from CT\cite{7RA-UNet}, and abnormal lymph nodes segmentation from PET/CT\cite{8AAR} and etc. \par
Although powerful representation learning capabilities, local translation invariance and filter sharing properties have made CNN-based approaches the de facto selection for image segmentation, they still have their own limitations. For instance, the insufficient capability to capture explicit global context and long-range relations owing to the intrinsic locality of convolution operations. Some studies tried to employ dilated convolution\cite{4deeplab}, image pyramids\cite{9pyramid}, prior-guided\cite{6bpd, 10Shape-Aware, 11wan2020superbpd}, multi-scale fusion\cite{12wang2020deep,13cheng2020higherhrnet}, and self-attention mechanisms\cite{14att-unet,15li2018pyramid} based CNN features to address these limitations. However, these studies exist weakness to extract global context features in the task of medical image segmentation, especially for the objects that have large inter-patient variation in terms of shape, scale and texture. \par
Transformers\cite{16vaswani2017attention}, initially is proposed for sequence-to-sequence modeling in nature language processing (NLP) tasks, such as machine translation, sentiment analysis, information extraction, and etc. Recently, the vision transformer (ViT) architecture \cite{17vit,18touvron2021training,19graham2021levit}, which tries to apply transformer to vision tasks, has achieved state-of-the-art results for image classification via pre-training on the large-scale dataset. Later, Transformer-based architectures have also been studied for semantic segmentation, such as SETR\cite{20zheng2021rethinking}, Swin Transformer\cite{21swin}, Swin-UNet\cite{22cao2021swinunet}, TransUNet\cite{23trans-unet}. However, the main limitation of these Transformer-based methods lies in the high requirement of computation power, which impedes them to run in real-time applications, for example, radiotherapy. \par

% Inspired by the LeViT transformer\cite{19graham2021levit}, which is designed for fast inference image classification with hybrid transformer and convolution blocks, we proposed LeViT-UNet for 2D medical image segmentation in this paper. To the best of our knowledge, LeViT-UNet is the first work that studied the speed and accuracy with Transformer-Based architecture for the medical image segmentation task. A comparison of the speed and performance operated in various convolution-based and transformer-based methods for Synapse dataset in shown in Figure 1. \par

Recently, LeViT\cite{19graham2021levit} is proposed for fast inference image classification with hybrid transformer and convolution blocks, which optimizes the trade-off between accuracy and efficiency. However, this architecture has not fully leveraged various scales of feature maps from transformer and convolution blocks, which are conducive to image segmentation. Inspired by the LeViT, we propose LeViT-UNet for 2D medical image segmentation in this paper, which aims to make faster encoder with transformer and improve the segmentation performance. To the best of our knowledge, LeViT-UNet is the first work that studies the speed and accuracy with transformer-based architecture for the medical image segmentation task. A comparison of the speed and performance operated in various convolution-based and transformer-based methods for Synapse dataset in shown in Figure 1. We can see that the our LeViT-UNets achieve competitive performance compared the fast CNN-based models. Meanwhile, performance of LeViT-UNet-384 surpasses the previous state-of-the-art transformer-based method, such as TransUnet and Swin-UNet. \par

\begin{figure}[htbp]
\centering
\subfigure[DSC vs Speed]{
\begin{minipage}[t]{0.5\linewidth}
\centering
\includegraphics[height=8cm, width=8cm]{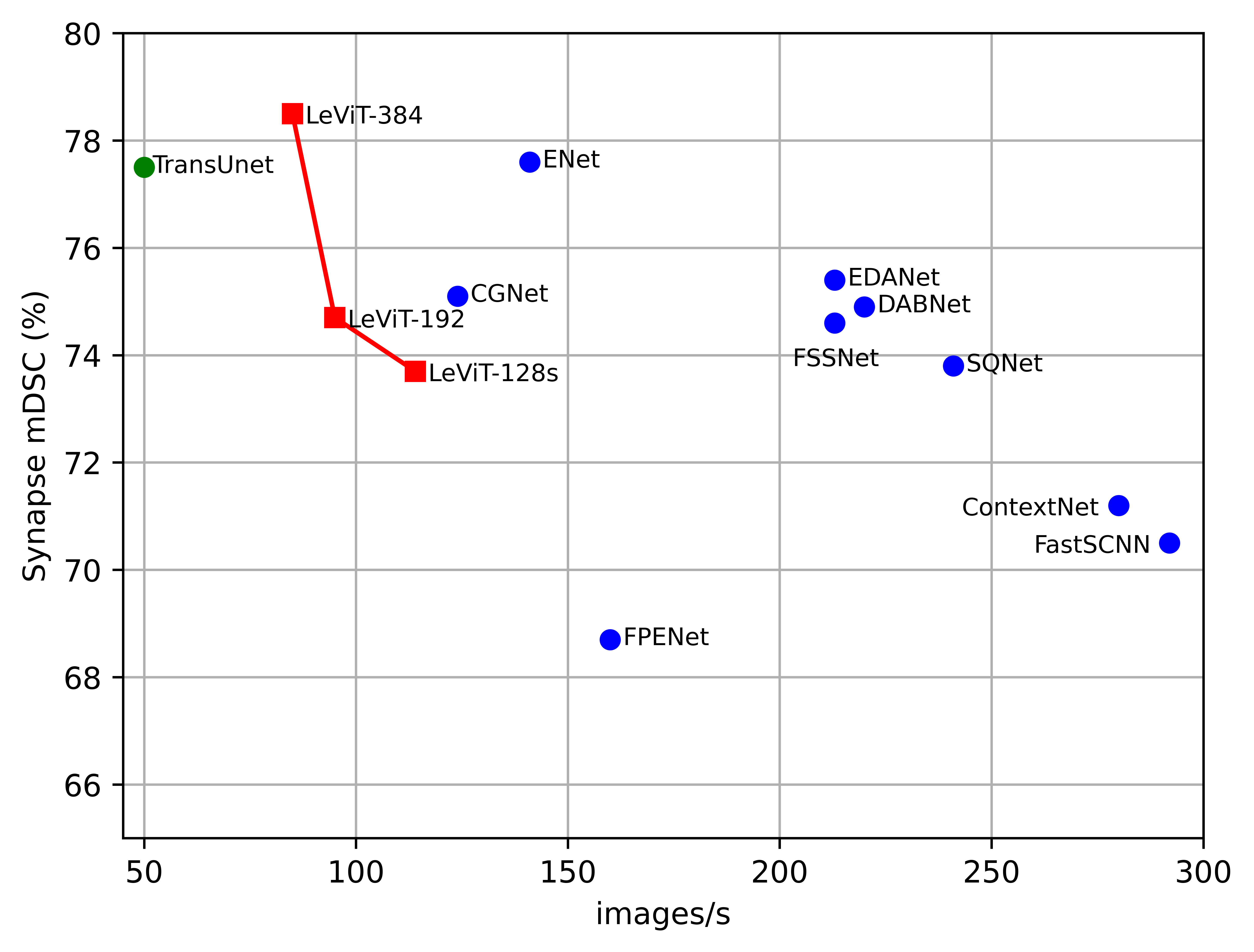} 
%\caption{fig1}
\end{minipage}%
}%
\subfigure[HD vs Speed]{
\begin{minipage}[t]{0.5\linewidth}
\centering
\includegraphics[height=8cm, width=8cm]{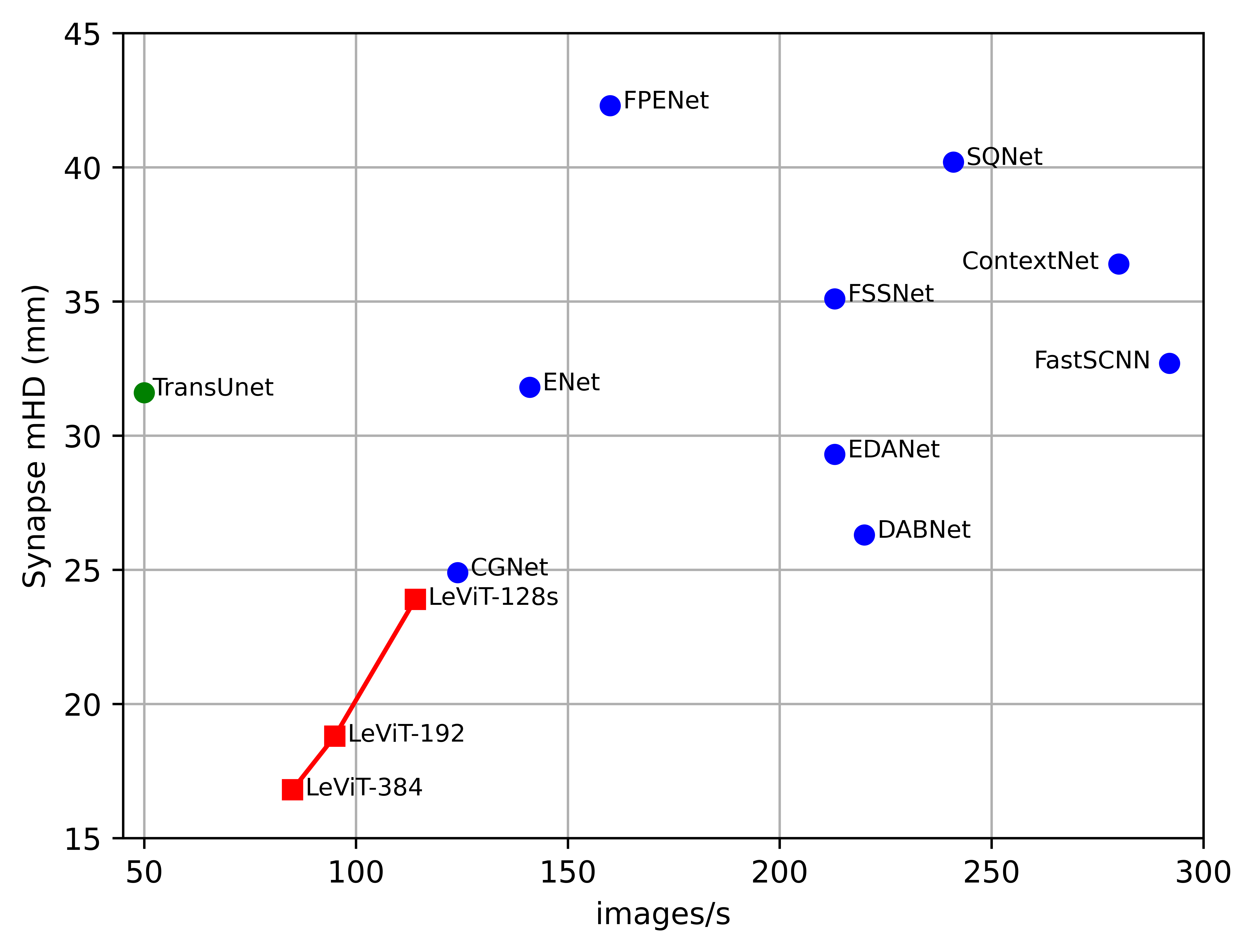} 
%\caption{fig2}
\end{minipage}%
}%
\centering
\caption{Speed and accuracy for convolution-based and visual transformers-based method, testing on Synapse dataset. Left: The speed and Dice similarity coefficient (DSC). Right: The speed and Hausdorff distance (HD).}
\end{figure}

The proposed LeViT-UNet consists of an encoder, a decoder and several skip connections. Here, the encoder is built based on LeViT transformer blocks, and the decoder is built based on convolution blocks. Motivated by the U-shape architecture design, the various resolution feature maps, which are extracted from the transformer blocks of LeViT is then upsampled, are concatenated and passed into decode blocks with skip connections. We find that such design could integrate the merits of the Transformer for global features extraction and the CNNs for local feature representation. Our experiments demonstrate that LeViT-UNet could improve both accuracy and efficiency of the medical image segmentation task. The main contributions of our work can be summarized as follows:
(1)	We propose a novel light-weight, fast and high accuracy transformer-based segmentation architecture, named LeViT-UNet, which integrates a multi-stage transformer block in the encoder with LeViT;
(2) We present a new method to fuse multi-scale feature maps extracted from the transformer and convolutional blocks, which could sufficiently integrate the global and local features in various scales;
(3)	Extensive experiments are conducted which demonstrated that the proposed method is competitive with other state-of-the-art methods in terms of accuracy and efficiency.  \par

\section{Related Works}
\label{sec:headings}
\paragraph{CNN-based methods:}
CNNs served as the standard network model have been extensively studied in medical image segmentation. The typical U-shaped network, U-Net\cite{2unet}, which consists of a symmetric encoder and decoder network with skip connections, has become the de-facto choice for medical image analysis. Afterwards, various U-Net like architectures is proposed, such as Res-UNet\cite{24res-unet}, Dense-UNet\cite{25denseunet}, V-Net\cite{26V-Net} and 3D-UNet\cite{27-3D-UNet}. While CNN-based methods have achieved much progress in medical image segmentation, they still cannot fully meet the clinical application requirements for segmentation accuracy and efficiency owing to its intrinsic locality of convolution operations and its complex data access patterns.
\paragraph{Self-attention mechanisms to complement CNNs:}
Several works have attempted to integrate self-attention mechanism into CNNs for segmentation. The main purpose is to catch the attention weight in terms of channel-wise or spatial shape. For instance, the squeeze-and-excite network built an attention-like module to extract the relationship between each feature map of a layer\cite{28senet}. The dual attention network appended two types of attention modules to model the semantic interdependencies in spatial and channel dimensions respectively\cite{29fu2019dual}. The Attention U-Net proposed an attention gate to suppress irrelevant regions of a feature map while highlighting salient features for segmentation task. Although these strategies could improve the performance of segmentation, the ability of extracting long-rang semantic information still need to be addressed.
\paragraph{Transformers:}Recently, Vision Transformer (ViT) achieved state-of-the-art on ImageNet classification by using transformer with pure self-attention to input images\cite{17vit}. Afterward, different ViT variants have been proposed, such as DeiT\cite{18touvron2021training}, Swin\cite{21swin}, and LeViT\cite{19graham2021levit}. Some works attempted to apply transformer structure to medical segmentation. For example, Medical Transformer (MedT) introduced the gated axial transformer layer into existing architecture. TransUNet\cite{23trans-unet} integrated the Transformers into U-Net, which utilized the advantage from both Transformers and CNN. Swin-UNet\cite{22cao2021swinunet} was proposed which employed pure transformer into the U-shaped encoder-decoder architecture for global semantic feature learning. In this paper, we attempt to apply LeViT transformer block as basic unit in the encoder of a U-shaped architecture, which trade-off the accuracy and efficiency for medical image segmentation. Our work will likely provide a benchmark comparison for the fast segmentation with Transformer in the field of medical image analysis.

\section{Method}
Given an input image of height (H) x width (W) x channel (C), the goal of the image segmentation task is to predict the corresponding pixel-wise label of H x W. Unlike the conventional UNet which employs convolutional operations to encode and decode features, we apply LeViT module in the encoder part to extract the features and keep the decoder part same as UNet. In the following part, we will introduce the overall LeViT-UNet architecture in Section 3.1. Then, the component of encoder and decoder in the LeViT-UNet will be elaborated in Section 3.2 and 3.3, respectively.

\subsection{The overall Architecture of LeViT-UNet}
The architecture of LeViT-UNet is present in Figure 2. It is composed of an encoder and a decoder. Here, we apply LeViT module in the encoder part to extract long-range structural information from the feature maps. The LeViT is a hybrid neural network which is composed of convnets and vision transformers.

\begin{figure} \centering  \includegraphics[height=8cm, width=12cm]{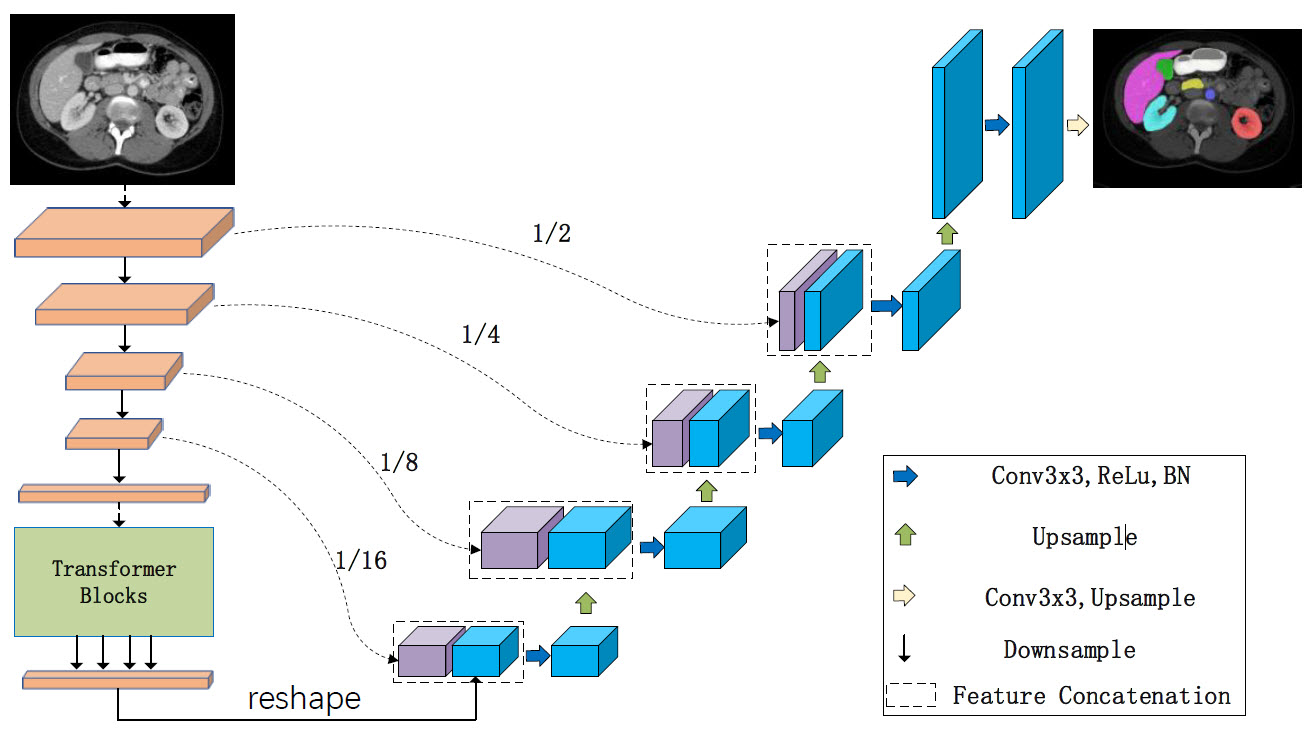}  \caption{The architecture of LeViT-UNet, which is composed of encoder (LeViT block), decoder and skip connection. Here, the encoder is constructed based on LeViT module.}  \end{figure}

\subsection{LeViT as Encoder}
Following \cite{19graham2021levit}, we apply LeViT architecture as the encoder, which consists of two main parts of components: convolutional blocks and transformer blocks. Specifically, there are 4 layers of 3x3 convolutions with stride 2 in the convolutional blocks, which could perform the resolution reduction. These feature maps will be fed into the transformer block, which could decrease the number of floating-point operations (FLOPs) that is known large in transformer blocks. Depending on the number of channels fed into the first transformer block, we design three types of LeViT encoder, which are named as LeViT-128s, LeViT-192 and LeViT-384, respectively. The block diagram of the LeViT-192 architecture is shown in Figure 3. Note that we concatenate the features from convolution layers and transformer blocks in the last stage of the encoder, which could fully leverage the local and global features in various scales.
\par

\begin{figure} \centering  \includegraphics[height=8cm, width=12cm]{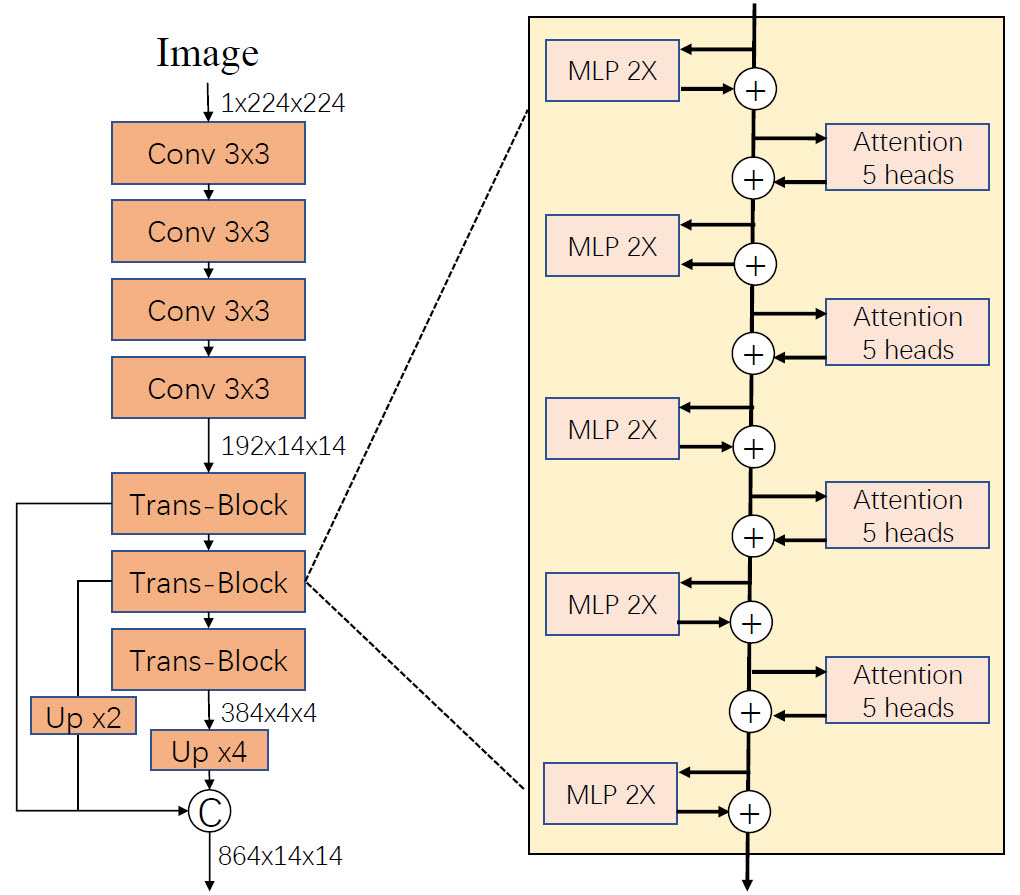}  \caption{Figure 3: Block diagram of LeViT-192 architecture. A sampling is applied before transformation in the second and third Trans-Block, respectively.}  \end{figure}

The transformer block can be formulated as: 
\begin{equation}
\hat{z}^n = {MLP}(BN(z^{n-1})) + z^{n-1}, 
\end{equation}

\begin{equation}
z^n = {MSA}(BN(\hat{z}^{n})) + \hat{z}^{n}, 
\end{equation}

Where $\hat{z}^n$ and $z^n$ represent the outputs of $MLP$ (Multiple Layer Perceptron) module and the $MSA$ (Multi-head Attention) module of the nth block, respectively. $BN$ means the batch normalization. Similar to the previous work [19], self-attention is computed as follows:

\begin{equation}
Attention(Q,K,V)\ =\ Softmax(\frac{QK^T}{\sqrt d}+B)V, 
\end{equation}
Where $Q$, $K$, $V$ are the query, key and value matrices, whose sizes are  $M^2$x$d$. $M^2$ and $d$ denote the number of patches and the dimension of the query or key. $B$ represents attention bias, which takes place of positional embedding and could provide positional information within each attention block.

\subsection{CNNs as Decoder}
Similar to U-Net, we concatenate the features from the decoder with skip connection. The cascaded upsampling strategy is used to recover the resolution from the previous layer using CNNs. For example, there are feature maps with the shape of H/16 x w/16 x D from the encoder. Then, we use cascaded multiple upsampling blocks for reach the full resolution of H x W, where each block consists of two 3x3 convolution layers, batch normalization layer, ReLU layer, and an upsampling layer. 

\section{Experiments and Results}
\subsection{Dataset}
\paragraph{Synapse multi-organ segmentation dataset (Synapse):}
This dataset includes 30 abdominal CT scans with 3779 axial contrast-enhanced abdominal clinical CT images in total. Following the splitting in [23], we use 18 cases for training and the other 12 cases for validation. We evaluated the performance with the average Dice Similarity Coefficient (DSC) and the average Hausdorff Distance (HD) on 8 abdominal organs, which are aorta, gallbladder, spleen, left kidney, right kidney, liver, pancreas, spleen, and stomach respectively. \par
\paragraph{Automated cardiac diagnosis challenge dataset (ACDC):}the ACDC dataset is collected from 150 patients using cine-MR scanners, splitting into 100 volumes with human annotations and the other 50 volumes which are private for the evaluation purpose. Here, we split the 100 annotated volumes into 80 training samples and 20 testing samples.

\subsection{Implementation details}
In this paper, we run all experiments based on Python 3.8, PyTorch 1.8.0 and Ubuntu 18.04.1 LTS. For all training samples, we apply augmentation strategies such as random flipping and rotations to increase data diversity. All trainings are performed on images size of 224x224 by using a Nvidia 3090 GPU with 24GB memory. For all models are trained by Adam optimizer with learning rate 1e-5 and weight decay of 1e-4. The cross entropy and Dice loss are use as objective function. We set input resolution 224x224, and batch size of 8. The training epochs are set as 350 and 400 for Synapse and ACDC dataset respectively. All transformer backbones in the LeViT were pretrained on ImageNet-1k to initialize the model parameters. Following [23], all 3D volume datasets are trained by slice and the predicted 2D slice are stacked together to build 3D prediction for evaluation.

\subsection{Experiment results on Synapse dataset}
We perform experiments with other state-of-the-art (SOTA) methods in terms of accuracy and efficiency as the benchmark for comparison with LeViT-UNet. Three variants of LeViT-UNet were designed. We identify them by the number of channels input to the first transformer block: LeViT-UNet-128s, LeViT-UNet-192, and LeViT-UNet-384, respectively. Following to [22][23], we report the average DSC and HD to evaluate our method on this dataset to demonstrate the generalization ability and robustness of our proposed method.

\subsubsection{Compare state-of-the-art methods}
The comparison of the proposed LeViT-UNet with other SOTA methods on the Synapse multi-organ CT dataset can be observed in Table 1. Experimental results show that LeViT-UNet-384 achieves the best performance in terms of average HD with 16.84 mm, which is improved by about 14.8 mm and 4.7 mm comparing the recently SOTA methods. It indicates that our approach can obtain better edge predictions. Comparing the transformer-based method, like TransUNet and SwinUNet, and other convolution-based method, like U-Net and Att-UNet, our approach still could achieve the competition result in terms of DSC. \par
The segmentation results of different methods on the Synapse dataset are shown in the Figure 4. We can see that the other three methods are more likely to under-segment or over segment the organs, for example, the stomach is under-segmented by TransUNet and DeepLabV3+ (as indicated by the red arrow in the third panel of the upper row), and over-segmented by UNet (as indicated by the red arrow in the fourth panel of the second row). Moreover, results in the third row demonstrate that our LeViT-UNet outputs are relatively smoother than those from other methods, which indicates that our method has more advantageous in boundary prediction. \par

\begin{figure} \centering  \includegraphics[height=8cm, width=12cm]{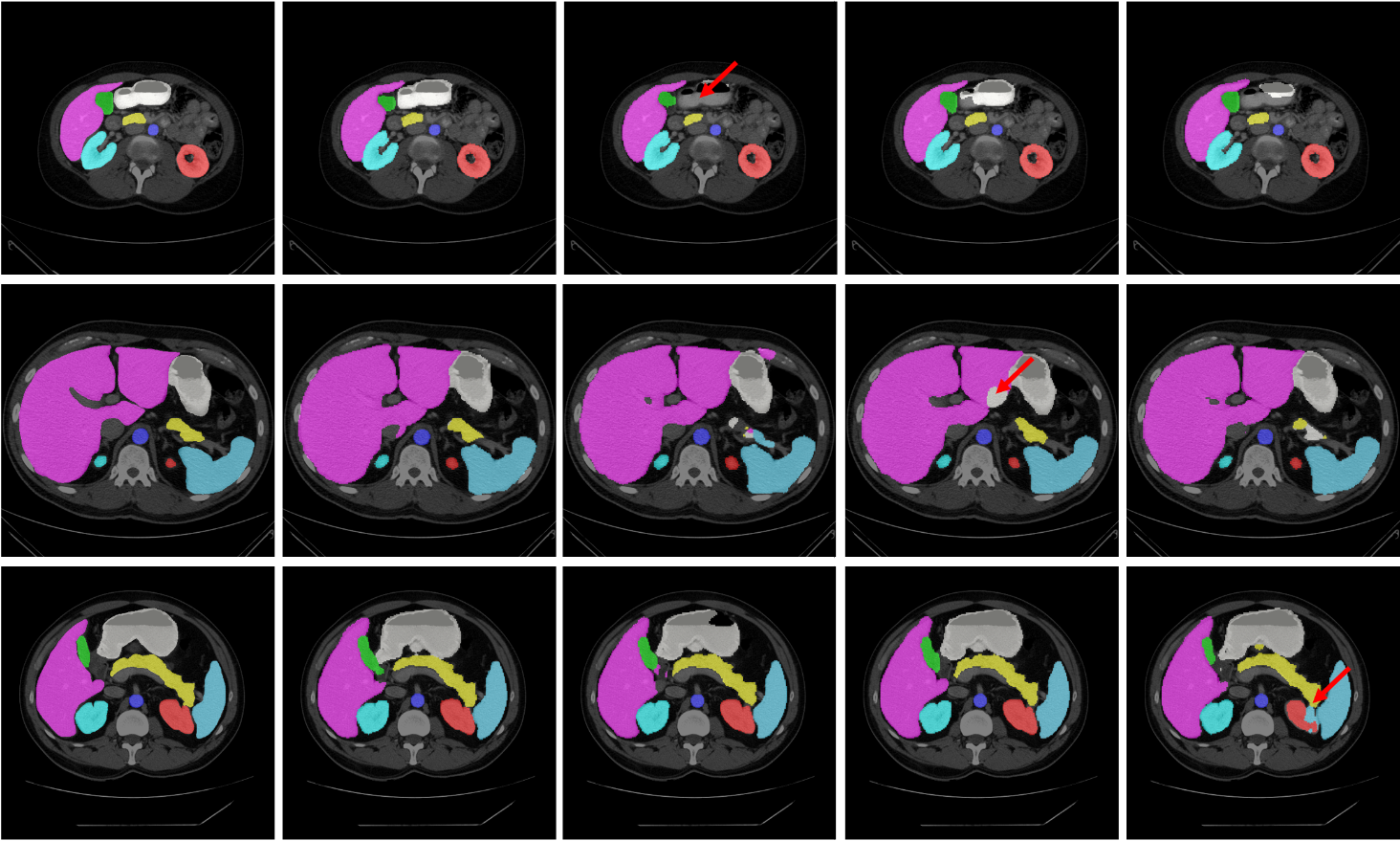}  \caption{Figure 4: Qualitative comparison of various methods by visualization From Left to right: Ground Truth, LeViT-UNet-384, TransUNet, UNet, and DeepLabv3+.}  \end{figure}

%\vspace{-1em}

%%%%%%%%%%%%%%%%%%%%%%%%%%%
\begin{table*}[t!]
\centering
\caption{Segmentation accuracy (average DSC\% and average HD in mm, and DSC for each organ) of different methods on the Synapse multi-organ CT dataset.}
\footnotesize
\resizebox{\textwidth}{!}{
\begin{tabular}{c|cc|cccccccc}
\hline
Methods &  DSC$\uparrow$ &HD$\downarrow$ & Aorta& Gallbladder& Kidney(L)& Kidney(R)& Liver& Pancreas& Spleen& Stomach\\
\hline
V-Net~\cite{26V-Net} &  68.81 & -& 75.34&51.87&77.10&\textbf{80.75}&87.84&40.05&80.56&56.98\\
DARR~\cite{30domain} &   69.77 & -& 74.74&53.77&72.31&73.24&94.08&54.18&89.90&45.96\\
U-Net~\cite{2unet} & 76.85 & 39.70& 89.07&\textbf{69.72}&77.77&68.60&93.43&53.98&86.67&75.58 \\
R50 U-Net~\cite{23trans-unet} & 74.68 & 36.87& 87.74&63.66&80.60&78.19&93.74&56.90&85.87&74.16 \\
R50 Att-UNet~\cite{23trans-unet}  & 75.57&  36.97& 55.92&63.91&79.20&72.71&93.56&49.37&87.19&74.95 \\
Att-UNet~\cite{14att-unet} & 77.77 & 36.02& \textbf{89.55}&68.88&77.98&71.11&93.57&58.04&87.30&75.75 \\
R50-Deeplabv3+~\cite{4deeplab} & 75.73  & 26.93  & 86.18  & 60.42  & 81.18  & 75.27  & 92.86  & 51.06  & 88.69  & 70.19  \\

R50 ViT~\cite{23trans-unet} & 71.29 &  32.87& 73.73&55.13&75.80&72.20&91.51&45.99&81.99&73.95\\
TransUnet~\cite{23trans-unet} & 77.48 &  31.69&87.23&63.13&81.87&77.02&94.08&55.86&85.08&75.62 \\
SwinUnet~\cite{21swin} & \textbf{79.13} & 21.55 &85.47&66.53&83.28&79.61&\textbf{94.29}&56.58&\textbf{90.66}&\textbf{76.60}\\
\hline
LeVit-UNet-128s & 73.69  & 23.92  & 86.45  & 66.13  & 79.32  & 73.56  & 91.85  & 49.25  & 79.29  & 63.70  \\ 
LeVit-UNet-192 & 74.67  & 18.86  & 85.69  & 57.37  & 79.08  & 75.90  & 92.05  & 53.53  & 83.11  & 70.61  \\ 
LeVit-Unet-384 & 78.53  & \textbf{16.84}  & 87.33  & 62.23  & \textbf{84.61}  & 80.25  & 93.11  & \textbf{59.07}  & 88.86  & 72.76  \\ 

\hline
\end{tabular}
}
\label{synapse}
\end{table*}

\subsubsection{Compare with fast segmentation methods}
Firstly, it can be seen that LeViT-UNet-384 achieves 78.53\% mDSC and 16.84mm mHD, which is the best among all methods in Table 2.  Particularly, we can find our proposed method is much faster than TransUNet, which integrates Transformer block into CNN. Then, to demonstrate the performance of accuracy-efficiency, we compare LeViT-UNet with other fast segmentation methods, such as ENet, FSSNet, FastSCNN and etc. In terms of the amount of parameters, our method is still needed to be improved, comparing to other fast segmentation methods,  like CGNet, ContextNet and ENet. However, our method has much fewer parameters than TransUNet. Moreover, we evaluate the runtime at different methods. Here, ENet (114 fps) and FPENet (160 fps) are slightly faster than LeViT-UNet-128s (114 fps), yet the HD are still needed to improve. Therefore, we conclude that LeViT-UNet is competitive with the current pure CNN efficient segmentation method with better performance. \par

%%%%%%%%%%%%table 2
\begin{table*}[t!]
\centering
\caption{Mean DSC and HD of the proposed LeViT-UNet compared to other state-of-the-art semantic segmentation methods on the Synapse dataset in terms of parameters and inference speed by FPS (frame per second). Number of parameters are listed in millions.}
\footnotesize
\resizebox{\textwidth}{!}{
\begin{tabular}{c|cc|cccccccc|ccc}
\hline
Methods &  DSC$\uparrow$ &HD$\downarrow$ & Aorta& Gallbladder& Kidney(L)& Kidney(R)& Liver& Pancreas& Spleen& Stomach & \# params(M) & FLOPs(G) & FPS \\
\hline
CGNet~\cite{cgnet} & 75.08  & 24.99  & 83.48  & 65.32  & 77.91  & 72.04  & 91.92  & 57.37  & 85.47  & 67.15  & 0.49  & 0.66  & 124  \\ 
ContextNet~\cite{contextnet} & 71.17  & 36.41  & 79.92  & 51.17  & 77.58  & 72.04  & 91.74  & 43.78  & 86.65  & 66.51  & 0.87  & 0.16  & 280  \\ 
DABNet~\cite{dabnet} & 74.91  & 26.39  & 85.01  & 56.89  & 77.84  & 72.45  & 93.05  & 54.39  & 88.23  & 71.45  & 0.75  & 0.99  & 221  \\ 
EDANet~\cite{edanet} & 75.43  & 29.31  & 84.35  & 62.31  & 76.16  & 71.65  & 93.20  & 53.19  & 85.47  & 77.12  & 0.69  & 0.85  & 213  \\ 
ENet~\cite{enet} & 77.63  & 31.83  & 85.13  & 64.91  & 81.10  & 77.26  & 93.37  & 57.83  & 87.03  & 74.41  & 0.36  & 0.50  & 141  \\ 
FPENet~\cite{fpenet} & 68.67  & 42.39  & 78.98  & 56.35  & 74.54  & 64.36  & 90.86  & 40.60  & 78.30  & 65.35  & 0.11  & 0.14  & 160  \\ 
FSSNet~\cite{fssnet} & 74.59  & 35.16  & 82.87  & 64.06  & 78.03  & 69.63  & 92.52  & 53.10  & 85.65  & 70.86  & 0.17  & 0.33  & 213  \\ 
SQNet~\cite{sqnet} & 73.76  & 40.29  & 83.55  & 61.17  & 76.87  & 69.40  & 91.53  & 56.55  & 85.82  & 65.24  & 16.25  & 18.47  & 241  \\ 
FastSCNN~\cite{fastscnn} & 70.53  & 32.79  & 77.79  & 55.96  & 73.61  & 67.38  & 91.68  & 44.54  & 84.51  & 68.76  & 1.14  & 0.16  & \textbf{292}  \\ 
TransUNet~\cite{23trans-unet} & 77.48  & 31.69  & 87.23  & 63.13  & 81.87  & 77.02  & 94.08  & 55.86  & 85.08  & 75.62  & 105.28  & 24.64  & 50  \\ 
\hline
LeViT-UNet-128s & 73.69  & 23.92  & 86.45  & 66.13  & 79.32  & 73.56  & 91.85  & 49.25  & 79.29  & 63.70  & 15.91  & 17.55  & 114  \\ 
LeViT-UNet-192 & 74.67  & 18.86  & 85.69  & 57.37  & 79.08  & 75.90  & 92.05  & 53.53  & 83.11  & 70.61  & 19.90  & 18.92  & 95  \\ 
LeViT-UNet-384 & \textbf{78.53}  & \textbf{16.84}  & 87.33  & 62.23  & 84.61  & 80.25  & 93.11  & 59.07  & 88.86  & 72.76  & 52.17  & 25.55  & 85  \\ 
\hline
\end{tabular}
}
\label{synapse_2}
\end{table*}

\subsubsection{Ablation study}
%\vspace{1ex}\noindent\textbf{Effect of the number of transformer blocks:}
We conduct a variety of ablation studies to thoroughly evaluate the proposed LeViT-UNet architecture and validate the performance under different settings, including: 1) without and with transformer blocks; 2) the number of skip-connections; 3) without and with pretraining.

%%%%%%%%%%%%table 3
\begin{table*}[t!]
\centering
\caption{Ablation study w/o Transformer blocks.}
\footnotesize
\resizebox{\textwidth}{!}{
\begin{tabular}{c|cc|cccccccc|cc}
\hline
Methods &  DSC$\uparrow$ &HD$\downarrow$ & Aorta& Gallbladder& Kidney(L)& Kidney(R)& Liver& Pancreas& Spleen& Stomach & \# params(M) & FLOPs(G) \\
\hline
LeViT-UNet-128s-Conv & 72.44  & 41.63  & 84.29  & 59.11  & 77.70  & 69.20  & 91.93  & 44.18  & 87.60  & 65.52  & 5.46  & 14.54  \\ 
LeViT-UNet-192-Conv & 74.42  & 35.41  & 85.34  & 62.90  & 81.39  & 72.80  & 91.76  & 44.95  & 88.84  & 67.36  & 5.97  & 15.30  \\ 
LeViT-UNet-384-Conv & 74.59  & 30.19  & 85.49  & 62.52  & 83.00  & 73.87  & 91.91  & 43.47  & 88.75  & 67.69  & 11.94  & 17.70  \\ 
\hline
LeViT-UNet-128s & 73.69  & 23.92  & 86.45  & \textbf{66.13}  & 79.32  & 73.56  & 91.85  & 49.25  & 79.29  & 63.70  & 15.91  & 17.55  \\ 
LeViT-UNet-192 & 74.67  & 18.86  & 85.69  & 57.37  & 79.08  & 75.90  & 92.05  & 53.53  & 83.11  & 70.61  & 19.90  & 18.92  \\ 
LeViT-UNet-384 &\textbf{78.53}  & \textbf{16.84}  & \textbf{87.33}  & 62.23  & \textbf{84.61}  & \textbf{80.25}  & \textbf{93.11}  & \textbf{59.07}  & \textbf{88.86}  & \textbf{72.76}  & 52.17  & 25.55  \\ 
\hline
\end{tabular}
}
\label{synapse_3}
\end{table*}

\vspace{1ex}\noindent\textbf{Effect of the number of transformer blocks:}
Here, we compare the performance when Transformer blocks are utilized or not. We can see that adding transformer blocks leads to a better segmentation performance in terms of DSC and HD in the Table 3. These results show that the transformer block could improve performance owing to its innate global self-attention mechanisms. Moreover, the channel number of feature maps that input to the transformer block could improve the HD performance significantly. It reduced the HD about 7.08mm and 11.44mm with/ without transformer blocks respectively from the channel number of 128 to 384. Meanwhile, we find that the number of channels gives more influence on the LeViT-UNet method than LeViT-UNet, which did not include transformer blocks. It can be seen that the DSC is boosted to 1.25\%, 0.25\%, and 4.84\% with transformer blocks, respectively. Particularly, the performance of HD is improved to 17.71mm, 16.55 and 13.35 from LeViT-UNet 128s to LeViT-UNet-384, respectively\par

\vspace{1ex}\noindent\textbf{Effect of the number of skip connections:} We investigate the influence of skip-connections on LeViT-UNet. The results can be seen in Table 4. Note that “1-skip” setting means that we only apply one time of skip-connection at the 1/2 resolution scale, and “2-skip”, “3-skip” and “4-skip” are inserting skip-connections at 1/4, 1/8 and 1/16, respectively. We can find that adding more skip-connections could result in better performance. Moreover, the performance gain of smaller organs, like aorta, gallbladder, kidneys, is more obvious than that of larger organs, like liver, spleen and stomach.

%%%%%%%%%%%%table 4
\begin{table*}[t!]
\centering
\caption{Ablation study on the number of skip-connection in LeViT-UNet. ( '\_N' means the number of skip connections)}
\footnotesize
\resizebox{\textwidth}{!}{
\begin{tabular}{c|cc|cccccccc}
\hline
Number of skip-connection &  DSC$\uparrow$ &HD$\downarrow$ & Aorta& Gallbladder& Kidney(L)& Kidney(R)& Liver& Pancreas& Spleen& Stomach   \\
\hline
LeViT-UNet-384\_N0 & 67.190  & 27.887  & 73.700  & 47.080  & 69.850  & 65.030  & 89.920  & 45.530  & 82.220  & 64.180  \\ 
LeViT-UNet-384\_N1 & 68.720  & 27.973  & 73.590  & 48.730  & 75.050  & 67.960  & 91.150  & 45.030  & 84.130  & 64.090  \\ 
LeViT-UNet-384\_N2 & 74.540  & 25.845  & 84.980  & 59.270  & 75.430  & 69.160  & 92.530  & 57.200  & 87.180  & 70.580  \\ 
LeViT-UNet-384\_N3 & 76.910  & 20.866  & 86.890  & 61.010  & 81.570  & 76.180  & 92.860  & 56.000  & 87.620  & 73.190  \\ 
LeViT-UNet-384\_N4 & 78.530  & 16.838  & 87.330  & 62.230  & 84.610  & 80.250  & 93.110  & 59.070  & 88.860  & 72.760  \\ 
\hline
\end{tabular}
}
\label{synapse_4}
\end{table*}

\vspace{1ex}\noindent\textbf{Effect of pre-training:} The pre-training affected the performance of Transformer-based models, which can be attributed that they do not have an inductive bias to focus on nearby image elements\cite{19graham2021levit}. Hence, a large dataset is needed to regularize the model. Interestingly, we found that pre-training did not cause much influence of performance with LeViT-UNet, especially on the evaluation of DSC. We can see that the DSC is higher without pre-training by the LeViT-UNet-128s and LeViT-UNet-192. However, as the LeViT-UNet-384, we found that the pre-training is helpful to improve the performance. It indicated that the pre-training causes much influence to the transformer-based model which have larger parameters, like LeViT-UNet-384, which has about 52.1 million parameters, in contrast with 15.9 million and 19.9 million parameters in LeViT-UNet-128s and LeViT-UNet-192, respectively.

%%%%%%%%%%%%table 5
\begin{table*}[t!]
\centering
\caption{Ablation study of influence of pretrained strategy. ('-N' means without pretraining on ImageNet)}
\footnotesize
\resizebox{\textwidth}{!}{
\begin{tabular}{c|cc|cccccccc}
\hline
Methods &  DSC$\uparrow$ &HD$\downarrow$ & Aorta& Gallbladder& Kidney(L)& Kidney(R)& Liver& Pancreas& Spleen& Stomach   \\
\hline
LeViT-UNet-128s-N & 76.30  & 23.77  & 85.49  & 62.91  & 82.74  & 75.57  & 93.03  & 54.76  & 87.16  & 68.75  \\ 
LeViT-UNet-192-N & 76.88  & 23.44  & 86.91  & 66.24  & 82.80  & 75.75  & 92.89  & 50.81  & 88.61  & 71.01  \\ 
LeViT-UNet-384-N & 77.98  & 23.69  & 86.05  & 65.99  & 82.89  & 76.99  & 93.24  & 58.01  & 89.78  & 70.89  \\ 
\hline
LeViT-UNet-128s & 73.69  & 23.92  & 86.45  & 66.13  & 79.32  & 73.56  & 91.85  & 49.25  & 79.29  & 63.70  \\ 
LeViT-UNet-192 & 74.67  & 18.86  & 85.69  & 57.37  & 79.08  & 75.90  & 92.05  & 53.53  & 83.11  & 70.61  \\ 
LeViT-UNet-384 & 78.53  & 16.84  & 87.33  & 62.23  & 84.61  & 80.25  & 93.11  & 59.07  & 88.86  & 72.76  \\
\hline
\end{tabular}
}
\label{synapse_5}
\end{table*}

\subsection{Experiment results on ACDC dataset}
To demonstrate the generalization ability of LeViT-UNet, we train our model on ACDC MR dataset for automated cardiac segmentation. We can observe that our proposed LeViT-UNet could achieve the better results in terms of DSC in the Table 6. Compared with Swin-UNet[22] and TransUNet[23], we can see that our LeViT-UNet achieve comparable DSC; for instance, the LeViT-UNet-192 and LeViT-Unet-384 achieve 90.08\% and 90.32\% DSC.

%%%%%%%%%%%%table 6
\begin{table*}[t!]
\centering
\caption{Segmentation performance of different methods on the ACDC dataset.}
\begin{tabular}{c|c|ccc}
\hline
Methods & DSC$\uparrow$ & RV & Myo& LV\\
\hline
R50 U-Net & 87.55 & 87.10&80.63& 94.92\\
R50 Att-UNet  & 86.75& 87.58&79.20&93.47 \\
R50 ViT & 87.57&86.07&81.88&94.75\\
TransUnet & 89.71&88.86&84.53&95.73 \\
SwinUnet & 90.00 &88.55&85.62&95.83\\
\hline
LeViT-UNet-128s & 89.39  & 88.16  & \textbf{86.97}  & 93.05  \\ 
LeViT-UNet-192 & 90.08  & 88.86  & 87.50  & \textbf{93.87}  \\
LeViT-UNet-384 & \textbf{90.32}  & \textbf{89.55}  & 87.64  & 93.76  \\
\hline
\end{tabular}
\label{acdc}
\end{table*}

\subsection{Discussion}
In this work, we apply LeViT as theencoder into UNet architecture. The feature maps from three Transformer blocks are directly concatenated after upsampling. In the future work, we will explore the ways to fuse multi-scale global feature maps from Transformer blocks. Moreover, the resolution of input image is down-scaled to 1/16 before the Transformer blocks in order to reduce the computation complexity, which may have effect on the performance of segmentation. We expect to design more efficient architectures that could keep that balance between the speed and the accuracy by using Transformer-based methods. Lastly, we would like to explore the applications of LeViT-UNet in 3D medical image segmentation.

\section{Conclusion}
Transformers are good at modeling long-range dependency with self-attention mechanism. In this paper, we present the first study that integrate LeViT into UNet-like architecture for the general medical image segmentation task. The proposed LeViT-UNet makes fully leverage of the advantage of Transformers to build strong global context while keeping the merit of CNN to extract low-level features. Extensive experiments demonstrate that compared with current SOTA methods, the proposed LeViT-UNet has superior performance and good generalization ability. Moreover, the proposed LeViT-UNet shows the ability of trade-off between accuracy and efficiency. In the future, we’d like to optimize further the structure of LeViT-UNet, which could compete with other CNN-based fast segmentation methods. \par

%\section{Reference}
\bibliographystyle{IEEEtran}

\bibliography{reference}

\end{document}